\def\year{2018}\relax
\documentclass[letterpaper]{article} 
\usepackage{aaai18}  
\usepackage{times}  
\usepackage{helvet}  
\usepackage{courier}  
\usepackage{url}  
\usepackage{graphicx}  
\usepackage{bbding}

\usepackage{booktabs}       
\usepackage{amsmath}
\usepackage{amsfonts}
\usepackage{bm}
\usepackage{algorithm}
\usepackage{algorithmic}
\usepackage{multirow}
\usepackage{subfigure}
\usepackage{fancyhdr}

\usepackage{amssymb}
\usepackage{amsthm}
\newtheorem{thm}{Theorem}

\newtheorem{lemma}{Lemma}
\newtheorem{defi}{Definition}
\newtheorem{remark}{Remark}
\newtheorem{condition}{Condition}

\begin{document}
%
\title{Proximal Alternating Direction Network: \\A Globally Converged Deep Unrolling Framework}
\author{Risheng Liu$^{1,2,}$\thanks{Corresponding Author.}, \ Xin Fan$^{1,2}$, \ Shichao Cheng$^{1,2,3}$, \ Xiangyu Wang$^{1,2}$, \ Zhongxuan Luo$^{1,2,3}$\\
$^1$DUT-RU International School of Information Science \& Engineering, Dalian University of Technology, Dalian, China\\
$^2$Key Laboratory for Ubiquitous Network and Service Software of Liaoning Province, Dalian, China\\
$^3$School of Mathematical Science, Dalian University of Technology, Dalian, China\\
\{rsliu, xin.fan, zxluo\}@dlut.edu.cn, \  shichao.cheng@outlook.com, \ wxy9326@gmail.com
}
\maketitle
\thispagestyle{fancy}          
\fancyhead{}                      
\lhead{Thirty-second AAAI Conference on Artificial Intelligence (AAAI-2018)}
\renewcommand{\headrulewidth}{0pt}

\begin{abstract}
Deep learning models have gained great success in many real-world applications. However, most existing networks are typically designed in heuristic manners, thus lack of rigorous mathematical principles and derivations. Several recent studies build deep structures by unrolling a particular optimization model that involves task information. Unfortunately, due to the dynamic nature of network parameters, their resultant deep propagation networks do \emph{not} possess the nice convergence property as the original optimization scheme does. This paper provides a novel proximal unrolling framework to establish deep models by integrating experimentally verified network architectures and rich cues of the tasks. More importantly, we \emph{prove in theory} that 1) the propagation generated by our  unrolled deep model globally converges to a critical-point of a given variational energy, and 2) the proposed framework is still able to learn priors from training data to generate a convergent propagation even when task information is only partially available. Indeed, these theoretical results are the best we can ask for, unless stronger assumptions are enforced. Extensive experiments on various real-world applications verify the theoretical convergence and demonstrate the effectiveness of designed deep models.
\end{abstract}

\section{Introduction}

%
%


In last years, deep models (a.k.a. deep neural networks) have produced the state-of-the-art performance in many application fields, such as image processing, object recognition, natural language processing, and bioinformatics.
On the downside, these existing approaches are typically designed based on heuristic understandings of a particular problem, and trained using engineering experience to implement multi-layered feature propagations,~e.g., \cite{krizhevsky2012imagenet,simonyan2014very,he2016deep}. Therefore, they lack solid theoretical guidances
and interpretations. More importantly, it is challenging to incorporate the
mathematical rules and principles of the considered task into these existing networks.


Alternatively, several recent works, e.g., \cite{gregor2010learning,schmidt2014shrinkage,andrychowicz2016learning}, build their networks using a specific optimization model and iteration scheme. The main idea is to unroll numerical algorithms and constitute their network architectures based on the resulted iterations. In this way, these approaches successfully incorporate the information of a predefined energy into the network propagation. Nevertheless, due to the dynamical nature of parameterized iterations, existing theoretical results, especially convergence, from the optimization area are not valid at all. Furthermore, the unrolled deep models are often with limited flexibility and adaptability as the basic architectures are restricted by the particular iteration scheme.

To partially overcome limitations in existing approaches,
this work attempts to develop a simple, flexible and efficient framework to build deep models
for various real-word applications.
The launching point of our work is from Energy-Based Models (EBMs)~\cite{teh2003energy,zhao2016energy}. EBMs are a series of methods, which associate a scalar energy to each configuration of observations and their
interested perditions. The inference of EBM consists of searching for a configuration of variables that minimizes the energy.
In this work, we consider the following energy minimization formulation
\begin{equation}
\inf\limits_{\mathbf{x}}\left\{\mathcal{F}(\mathbf{x}):=f(\mathbf{x};\mathbf{y}) +   r(\mathbf{x})\right\}  \label{eq:model},
\end{equation}
where $\mathbf{y}$ and $\mathbf{x}$ are the observed and predicted variables, respectively, $r$ reveals the priors of predictions, and $f$ is a measure of compatibility, i.e., fidelity, between $\mathbf{x}$ and $\mathbf{y}$.

We establish a novel proximal
framework to unroll the general energy in Eq.~\eqref{eq:model}, and incorporate various experimentally efficient architectures into the resulted deep model. Promising theoretical properties and practical performance will be also demonstrated.
The main advantages of our proposed framework against existing optimization-unrolled deep models can be distilled to the following three points.

\textbf{Insensitive Unrolling Scheme:} Most existing iteration-unrolling based deep models are strictly confined to some special types of energy formulations. For example, architectures in \cite{schmidt2014shrinkage,chen2016trainable}
are deduced from the field-of-experts prior while networks in~\cite{gregor2010learning} are based on $\ell_1$-regularizations. In contrast, our unrolling strategy only depends on the separable structure and functional properties of $\mathcal{F}$, but is completely insensitive to particular forms of $f$ and $r$. We can even design deep models without knowing the form of $r$ so that our framework adapts to various challenging tasks and complex data distributions.

\textbf{Flexible Built-in Architecture:}
On the one hand, the architectures in existing unrolled networks are deduced from fixed iteration schemes,
thus lacking flexibility. On the other hand, it has been revealed in many practical applications that heuristic deep architectures, e.g., ReLU and Batch Normalization, 
are extremely efficient though absent of theoretic analysis. Our studies show that under some mild conditions, our deep model can incorporate most existing empirically successful network architectures (even built by means of engineering tricks). In other words, we indeed theoretically offer the flexibility for existing deep architectures while taking the advantage of their efficiency.

\textbf{Convergence Guarantee:}
A fundamental weakness underlying existing networks is the elusiveness of theoretical analysis.
Especially,
little to no attention has been paid to the convergence behaviors of deep models\footnote{
Notice that the concept of ``convergence'' in this paper is not only related to the propagation of network parameters in the training phase, but the outputs of network architectures for both training and test phases. That is,
we consider the output of the $t$-th basic architecture as the $t$-th element of a sequence, and then investigate the convergence on the resulting sequence.}.
The main reason is
that even building networks by unrolling converged optimization algorithms, the dynamic nature of their parameters
and heuristic architectures
would still break the convergent guarantee in original iteration scheme. Contrarily, this paper proves that our designed deep models do have nice convergence properties. That is, under some mild conditions, the sequence generated by the proposed proximal alternating direction networks (PADNet) can converge
to a critical-point of Eq.~\eqref{eq:model} with relatively simple priors\footnote{We will formally discuss details of relatively ``simple'' and ``complex'' priors in the following sections.}. Furthermore, our propagation guarantees at least fixed-point convergence when handling complex priors, e.g., only with partial task/data information.
We argue that \emph{these theoretical results are the best we can ask for, unless stronger assumptions are enforced.}




\section{Proximal Alternating Direction Network}\label{sec:padnet}

In this section, we develop an alternating direction type unrolling scheme to generate the propagation sequence (denoted as $\{\mathbf{x}^k\}$) based on the energy model in Eq.~\eqref{eq:model}. As shown in the following, rather than directly calculating $\mathbf{x}^{k+1}$ from $\mathbf{x}^k$, we would like to first design cascaded propagations of two auxiliary variables (denoted as $\mathbf{u}^k$ and $\mathbf{v}^k$) corresponding to the fidelity and prior of the task, respectively. The residual type deep architectures are then incorporated for subsequence updating. Finally, a novel proximal error correction process is designed to control our propagation.

\textbf{Alternating Direction Scheme:}
For each $\mathbf{x}^k$, we introduce the Moreau-Yosida regularization~\cite{parikh2014proximal,xu2016mm} of $\mathcal{F}$ with parameter $\mu^k$ and auxiliary variable $\mathbf{u}$ to obtain the following regularized energy model
\begin{equation}
\begin{array}{c}
\mathcal{M}_{\mathcal{F}}^{\mu^k}(\mathbf{x}^k)=\inf\limits_{\mathbf{u}}\left\{f_{\mathbf{x}^k}^{\mu^k}(\mathbf{u}) + r(\mathbf{u})\right\}, \\
\mbox{where} \quad f_{\mathbf{x}^k}^{\mu^k}(\mathbf{u}):=f(\mathbf{u})+\frac{\mu^k}{2}\|\mathbf{u}-\mathbf{x}^{k}\|^2. \label{eq:pmodel}
\end{array}
\end{equation}
Now the problem is temporarily reduced to calculate $\mathbf{u}$ based on $\mathbf{x}^k$.
One common inference strategy for $\mathbf{u}$ in Eq.~\eqref{eq:pmodel} is to
introduce another auxiliary variable $\mathbf{v}$
and a Lagrange multiplier $\bm{\lambda}$ and then
perform alternating minimizations to the corresponding augmented Lagrange function, resulting to the following iteration scheme
\begin{eqnarray}
&\mathbf{u}^{k+1}=\arg\min\limits_{\mathbf{u}}  f_{\mathbf{x}^k}^{\mu^k}(\mathbf{u})+ \frac{\rho^k}{2}\|\mathbf{u}-(\mathbf{v}^{k}-\bm{\lambda}^k)\|^2, \label{eq:update-u}\\
&\mathbf{v}^{k+1} \in \arg\min\limits_{\mathbf{v}}r(\mathbf{v})+ \frac{\rho^k}{2}\|\mathbf{v}-(\mathbf{u}^{k+1}+\bm{\lambda}^k)\|^2,\label{eq:update-v}\\
&\bm{\lambda}^{k+1} = \bm{\lambda}^k + (\mathbf{u}^{k+1}-\mathbf{v}^{k+1}),\label{eq:update-lambda}
\end{eqnarray}
where  $\rho^k$ is a penalty parameter.

In this way, we actually perform the well-known
Alternating Direction Method of Multiplier (ADMM)~\cite{parikh2014proximal,lin2011linearized}
for the Moreau-Yosida regularized approximation of the original energy in Eq.~\eqref{eq:model} at each iteration.

\textbf{Built-in Deep Architecture:}
We then show how to incorporate deep architectures into the above base iteration. Specifically, we consider a residual formulation to replace the subproblem in Eq.~\eqref{eq:update-v}. That is, we  define the propagation of $\mathbf{v}$ as\footnote{Formally, we should denote the output of $t$-th residual unit at $k$-th iteration as $\mathbf{v}_t^k$. But in this paragraph, we temporarily omit the superscript $k$ to simplify the presentation.}
\begin{equation}
\mathbf{v}_T=\mathcal{N}_{\alpha}(\mathbf{v}_0;\bm{\mathcal{W}}_T):=\mathbf{v}_0 - \alpha\left(\sum_{t=0}^{T-1}\mathcal{G}(\mathbf{v}_t; \mathcal{W}_t)\right),\label{eq:network}
\end{equation}
where $\bm{\mathcal{W}}_T=\{\mathcal{W}_t\}_{t=0}^{T-1}$ is the set of learnable parameters, $\alpha$ is a step-size parameter, $\mathcal{G}$ is the basic network unit, $(\mathbf{v}_0,\mathbf{v}_{T})$ are the input and output of $\mathcal{N}_{\alpha}$ (at $T$-th stage), respectively. Notice that standard training strategies can be directly adopted to optimize parameters of our basic architecture. If necessary, one may further jointly fine-tune parameters of the whole network after the design phase.

It is easy to check that the network in Eq.~\eqref{eq:network} actually \emph{recursively} performs coordinate descent steps (i.e., $\mathbf{v}_{t+1}=\mathbf{v}_t-\alpha\mathcal{G}(\mathbf{v}_t)$) to propagate $\mathbf{v}$. So from optimization viewpoint, we interpret $\mathcal{G}$ as a descent-direction-estimation architecture for the optimization of the subproblem in Eq.~\eqref{eq:update-v}. While in more challenging scenario (e.g., hard to define an explicit and solvable $r$ for this subproblem), we can still learn built-in propagation architectures from training data to obtain our desired solution.



\textbf{Proximal Error Correction:}
Now it is ready to give the formal updating scheme of $\mathbf{x}^k$.
We can see that built-in architectures in Eq.~\eqref{eq:network} actually do not exactly optimize the original energy in Eq.~\eqref{eq:model}.  So it is necessary to introduce an additional step to control
our propagation at each iteration. Specifically,
denote $\mathbf{v}^{k+1}$ as the output of built-in network in Eq.~\eqref{eq:network} at $k$-th iteration. Then we adopt a proximal-gradient-like scheme \cite{DBLP:journals/corr/WangLSS17} to formally update $\mathbf{x}^{k+1}$
\begin{equation}
\begin{array}{l}
\mathbf{x}^{k+1}\in \\\arg\min\limits_{\mathbf{x}}r(\mathbf{x})+\frac{1}{2}\|\mathbf{x}-(\mathbf{v}^{k+1}-\nabla f_{\mathbf{x}^k}^{\mu^k}(\mathbf{v}^{k+1}))\|^2 \\
:=\mathtt{prox}_{r}\left(\mathbf{v}^{k+1}-\nabla f_{\mathbf{x}^k}^{\mu^k}(\mathbf{v}^{k+1})\right),\label{eq:err-corect}
\end{array}
\end{equation}
where $\mathtt{prox}_r$ is Moreau's proximal operator of $r$.

Overall, our proposed deep model, called \textbf{P}roximal \textbf{A}lternating \textbf{D}irection \textbf{Net}work (PADNet), is summarized in Alg.~\ref{alg:padm-test}. Notice that we actually consider PADNet in two different scenarios, which can be categorized by properties of prior regularization $r$ in Eq.~\eqref{eq:model}. That is:
\begin{itemize}
	\item \emph{Simple priors}: $\mathtt{prox}_r$ can be computed in closed-form.
	\item \emph{Complex priors}: $\mathtt{prox}_r$ is intractable or $r$ is unknown.
\end{itemize}

We perform error correction (i.e., Step~\ref{step:correction} in Alg.~\ref{alg:padm-test}) in the first case (Explicit PADNet or EPADNet for short) but directly propagate the output of built-in networks (i.e., Step~\ref{setp:network} in Alg.~\ref{alg:padm-test}) in the second case (Implicit PADNet or IPADNet for short). Theoretical results for these two different scenarios will be respectively proved
in the next section.

\begin{algorithm}
	\caption{Proximal Alternating Direction Network}\label{alg:padm-test}
	\begin{algorithmic}[1]
		\REQUIRE $\mathbf{u}^0$, $\mathbf{v}^{0}$, $\mathbf{x}^0$, $\bm{\lambda}^0$ and necessary parameters.
		\FOR{$k=0,1,2,\cdots,K-1$}
		\STATE $\mathbf{u}^{k+1}=\arg\min_{\mathbf{u}}  f_{\mathbf{x}^k}^{\mu^k}(\mathbf{u})+ \frac{\rho^k}{2}\|\mathbf{u}-(\mathbf{v}^{k}-\bm{\lambda}^k)\|^2$.\\
		 (Preliminary Estimation)
		\STATE $\mathbf{v}^{k+1} = \mathcal{N}_{\alpha^k}(\mathbf{u}^{k+1}+\bm{\lambda}^k;\bm{\mathcal{W}}_{T_k}^k)$.  \\ (Network Propagation)\label{setp:network}
		\STATE $\mathbf{x}^{k+1}=\mathtt{prox}_{r}\left(\mathbf{v}^{k+1}-\nabla f_{\mathbf{x}^k}^{\mu^k}(\mathbf{v}^{k+1})\right)$.
		\\ (Error Correction)\label{step:correction}
		\STATE $\bm{\lambda}^{k+1} = \bm{\lambda}^k + (\mathbf{u}^{k+1}-\mathbf{v}^{k+1})$.\\
		(Dual Updating)
		\ENDFOR  	
	\end{algorithmic}
\end{algorithm}

\section{Learning with Convergence Guarantee}\label{sec:train}

%

In general, unrolling task-aware optimization schemes may incorporate rich domain-knowledge into the network structure. Unfortunately, the sequence generated by most existing unrolled deep models will no longer have convergence guarantee, even though nice theoretical results have been proved and verified for their original optimization schemes.

Fortunately, we in this work demonstrate that under some mild conditions, the propagation generated by our PADNet is globally converged\footnote{Notice that ``globally converged'' in this paper is in the sense that the whole sequence generated by our deep model is converged and this concept has been widely used in non-convex optimization~\cite{attouch2010proximal} society.}, even with built-in network architectures designed in heuristic manners.


%

\subsection{Convergence Behavior Analysis of PADNet}

To make our paper self-contained, some necessary definitions should be presented before the formal analysis. Indeed, these concepts have been widely known in variational analysis and optimization and one may refer to \cite{rockafellar2009variational,attouch2010proximal} and references therein for more details.
\begin{defi} We give necessary definitions, including proper and lower semi-continuous, coercive and  semi-algebraic.
	\begin{itemize}
		\item A function $r: \mathbb{R}^n\to(-\infty, +\infty]$ is said to be proper and lower semi-continuous if $\texttt{dom}(r)\neq \emptyset$, where $\texttt{dom}(r) :=\{\mathbf{x}\in\mathbb{R}^n: r(\mathbf{x})<+\infty\}$ and $\liminf_{\mathbf{x}\to\mathbf{y}}r(\mathbf{x})\geq r(\mathbf{y})$ at any point $\mathbf{y}\in\texttt{dom}(r)$.
		\item A function $\mathcal{F}$ is said to be coercive, if $\mathcal{F}$ is bounded from below and $\mathcal{F}\to\infty$ if $\|\mathbf{x}\|\to\infty$, where $\|\cdot\|$ is the $\ell_2$ norm.
		\item A subset $S$ of $\mathbb{R}^n$ is a real semi-algebraic set if there exist a finit number of real polynomial functions
		$g_{ij}, h_{ij}: \mathbb{R}^n\to\mathbb{R}$ such that
		\begin{equation}
		S = \bigcup\limits_{j=1}^p\bigcap\limits_{i=1}^q\left\{\mathbf{x}\in\mathbb{R}^n: g_{ij}(\mathbf{x})=0 \ \mbox{and} \ h_{ij}(\mathbf{x})<0\right\}.
		\end{equation}
		A function $r:\mathbb{R}^n\to(-\infty,\infty]$ is called semi-algebraic if its graph $\{(\mathbf{x},z)\in\mathbb{R}^{n+1}:
		r(\mathbf{x})=z\}$ is a semi-algebraic subset of $\mathbb{R}^{n+1}$.
	\end{itemize}
\end{defi}

\begin{remark}
Indeed, many functions arising in learning and vision areas, including $\ell_0$ norm, rational $\ell_p$
norms (i.e., $p=p_1/p_2$ with positive integers $p_1$ and $p_2$) used in our experimental part
and their finite sums or products, are all semi-algebraic.
\end{remark}

In the following, we first analyze PADNet for tasks with simple priors.
Specifically, given a variable $\mathbf{x}$, we estimate the discrepancy between it and the optimal solution of
Eq.~\eqref{eq:pmodel} by the function
\begin{equation}
\mathcal{E}^k(\mathbf{x}):=\mathbf{g}_{\mathbf{x}} + \nabla f_{\mathbf{x}^k}^{\mu^k}(\mathbf{x}), \ \mbox{where} \ \mathbf{g}_{\mathbf{x}}\in \partial r(\mathbf{x}).\label{eq:error}
\end{equation}
Here $\mathcal{E}^k$ is deduced based on the first-order optimality condition of Eq.~\eqref{eq:pmodel} at $k$-th iteration.
Then with the following simple error condition, we  prove in Theorem~\ref{thm:global-convergence} that the propagation of EPADNet indeed globally converges to a critical-point of Eq.~\eqref{eq:model}. Please refer to supplemental materials for necessary preliminaries and all proofs of the proposed theories in this paper.

\begin{condition}(Error Condition)\label{con:error} The error function (in Eq.~\eqref{eq:error}) at $k$-th iteration should satisfy
$\|\mathcal{E}^k(\mathbf{x}^{k+1})\|\leq C_{\mathcal{E}}\|\mathbf{x}^{k+1}-\mathbf{x}^k\|$,
where $C_{\mathcal{E}} >0$
is a universal constant.
\end{condition}

\begin{thm}(Critical-Point Convergence of Explicit PADNet)\label{thm:global-convergence}
	Let $f$ be continuous differential, $r$ be proper and lower semi-continuous and $\mathcal{F}$ be
	coercive. Then EPADNet converges to a critical point of Eq.~\eqref{eq:model} under Condition~\ref{con:error}. That is,
	$\{\mathbf{x}^k\}$ generated by EPADNet is a bounded sequence and its any cluster point $\mathbf{x}^*$ is a critical point of Eq.~\eqref{eq:model} (i.e., satisfying
	$0\in\partial\mathcal{F}(\mathbf{x}^*)$). Furthermore, if $\mathcal{F}$ is semi-algebraic, then
	$\{\mathbf{x}^k\}$ globally converges to a critical point of Eq.~\eqref{eq:model}.
\end{thm}

\begin{remark}
With the semi-algebraic property of $\mathcal{F}$, we can also obtain
convergence rate of EPADNet based on a particular desingularizing function $\phi(s)=\frac{c}{\theta}(s)^{\theta}$ with a constant $c>0$ and parameter $\theta\in(0,1]$ (defined in~\cite{chouzenoux2016block}). Specifically, the sequence converges after finite iterations
if $\theta=1$. The linear and sublinear rates can be obtained if choosing $\theta\in[1/2,1)$ and $\theta\in(0,1/2)$, respectively.
\end{remark}

It has been verified that broad class of functions arising in learning problems (even \emph{nonconvex} and \emph{nonsmooth}) satisfy assumptions in Theorem~\ref{thm:global-convergence}. For example, both $\ell_0$ norm and rational $\ell_p$ norm with $p > 0$ (i.e., $p=p_1/p_2$, $p_1$ and $p_2$ are positive integers) are proper, lower semi-continuous and semi-algebraic.

Based on above analysis, we propose a learning framework (summarized in Alg.~\ref{alg:padm-train}) to adaptively design and train globally converged deep models for different learning tasks.
\begin{algorithm}
	\caption{The Learning Framework of PADNet}
	\begin{algorithmic}[1]
		\REQUIRE $\mathbf{u}^0$, $\mathbf{v}^0$, $\mathbf{x}^0$, $\bm{\lambda}^0$, $\mathcal{G}$, $t_{\texttt{max}}\geq1$, $k_{\texttt{max}}\geq1$, $\epsilon> 0$, $C_{\mathcal{E}}>0$, $\{\mu^k|2C_{\mathcal{E}}<\mu^k <\infty\}$, $\{\rho^k|\rho^k=(\gamma)^k\rho^0, \rho^0>0, \gamma > 1\}$ ($(\gamma)^k$ denotes $k$-th power of $\gamma$) and $\{\alpha^k| \alpha^k=\sqrt{1/\rho^k}\}$.
		\FOR{$k=0,1,2,\cdots$}
		\STATE  $\mathbf{u}^{k+1}=\arg\min_{\mathbf{u}} f_{\mathbf{x}^k}^{\mu^k}(\mathbf{u})+ \frac{\rho^k}{2}\|\mathbf{u}-(\mathbf{v}^{k}-\bm{\lambda}^k)\|^2$.
		\FOR{$t=0,1,2,\cdots$}
		\STATE Train $\mathcal{N}_{\alpha^k}$ with $\bm{\mathcal{W}}_{0:t}^{k}=\{\bm{\mathcal{W}}_{0:t-1}^{k},\mathcal{W}_t^k\}$ (i.e., $t+1$ basic units).
		\STATE  $\mathbf{v}_{t+1}^{k+1} = \mathcal{N}_{\alpha^k}(\mathbf{u}^{k+1}+\bm{\lambda}^k;\bm{\mathcal{W}}_{0:t}^{k})$.\label{step:v-net}
		\STATE  $\mathbf{x}_{t+1}^{k+1}=\mathtt{prox}_{r}(\mathbf{v}_{t+1}^{k+1}-\nabla f_{\mathbf{x}^k}^{\mu^k}(\mathbf{v}_{t+1}^{k+1}))$. \label{step:error-chekck}
		\IF{$\|\mathcal{E}^k(\mathbf{x}_{t+1}^{k+1})\|\leq C_{\mathcal{E}}\|\mathbf{x}_{t+1}^{k+1}-\mathbf{x}^k\|$  or $t\geq t_{\texttt{max}}$}
		\STATE  $\mathbf{x}^{k+1}=\mathbf{x}_{t+1}^{k+1}$, $\mathbf{v}^{k+1}=\mathbf{v}_{t+1}^{k+1}$, $T_{k}=t$, break.\label{setp:v-updata}
		\ENDIF
		\ENDFOR
		\STATE $\bm{\lambda}^{k+1} = \bm{\lambda}^k + (\mathbf{u}^{k+1}-\mathbf{v}^{k+1})$.\label{step:lambda}
		\IF{$\frac{\|\mathbf{x}^{k+1}-\mathbf{x}^{k}\|}{\|\mathbf{x}^k\|}\leq \epsilon$ or $k\geq k_{\texttt{max}}$}
		\STATE $K=k$, break.
		\ENDIF
		\ENDFOR
	\end{algorithmic}
	\label{alg:padm-train}
\end{algorithm}

\begin{remark}
	Theorem~\ref{thm:global-convergence} together with Alg.~\ref{alg:padm-train} actually provides a flexible framework with solid theoretical guarantee for deep model design and
	we only need to check whether built-in networks satisfy Condition~\ref{con:error} during their design phase.
	Furthermore, in general, any architectures satisfying this condition (even designed in engineering manner)
	can be incorporated into our deep models.
\end{remark}

In contrast, when handling tasks with complex priors, neither error checking (i.e., Step~\ref{step:error-chekck} in Alg.~\ref{alg:padm-train}) during design and training  nor
error correction (i.e., Step~\ref{step:correction} in Alg.~\ref{alg:padm-test}) during test  will be performed.
Therefore, we cannot obtain the same convergence results
as that in Theorem~\ref{thm:global-convergence}. Fortunately, by enforcing another easily satisfied condition to built-in architectures, we would still prove a fixed-point convergence guarantee for IPADNet.
\begin{condition}(Architecture Condition)\label{con:net}
For any given input $\mathbf{v}$, the architecture $\mathcal{N}_{\alpha}$ should satisfy
$
\|\mathcal{N}_{\alpha}(\mathbf{v})-\mathbf{v}\| \leq  C_{\mathcal{N}}\alpha$,
	where $C_{\mathcal{N}}>0$ is a universal constant.
\end{condition}
Notice that this bound condition is relatively weak and we can check that most commonly used linear and nonlinear operations in existing deep networks satisfy it.

\begin{thm}(Fixed-Point Convergence of Implicit PADNet)\label{thm:fixed-point-convergence}
	Let $f$ be continuous differential with bounded gradients. Then IPADNet is converged under Condition~\ref{con:net}. That is,  $\{(\mathbf{u}^k,\mathbf{v}^k,\bm{\lambda}^k)\}$ generated by IPADNet is a Cauchy sequence, so that is globally converged to a fixed-point.
\end{thm}

\begin{remark}\label{remark:fixed-point}
Theorem~\ref{thm:fixed-point-convergence} actually provides a theoretically guaranteed paradigm to fuse both analytical and empirical informations to build deep models for challenging learning tasks. That is to say, we can simultaneously design model-based fidelity function $f$ to reveal our theoretical understandings of the problem and learn complex priors from training data by model-free network architecture $\mathcal{N}_{\alpha}$.
\end{remark}

To end our analysis, we emphasize that the above convergence results
are the best we can ask for unless other stronger assumptions are made on the given learning task.

\subsection{Implementable Error Calculation}
It can be observed in Eq~\eqref{eq:error} that directly calculating $\mathcal{E}^k$ using its theoretical definition is challenging due to the subgradient term $\mathbf{g}_{\mathbf{x}}$. So we provide a calculable formulation based on the following derivations. Specifically, using Eq.~\eqref{eq:err-corect}, we have
\begin{equation}
\begin{array}{l}
\mathbf{x}^{k+1}=\mathtt{prox}_{r}\left(\mathbf{v}^{k+1}-\nabla f_{\mathbf{x}^k}^{\mu^k}(\mathbf{v}^{k+1})\right)\\
=\mathtt{prox}_{r} \Big(  \mathbf{x}^{k+1}-\nabla f_{\mathbf{x}^k}^{\mu^k}(\mathbf{x}^{k+1})  +(\mu^k-1)(\mathbf{x}^{k+1}\\
\quad  -\mathbf{v}^{k+1})-  \nabla f(\mathbf{v}^{k+1}) + \nabla  f(\mathbf{x}^{k+1}) \Big).
\end{array}\label{eq:error-prac}
\end{equation}
By setting $\mathbf{e}^{k+1}=(\mu^k-1)(\mathbf{x}^{k+1}-\mathbf{v}^{k+1})-\nabla f(\mathbf{v}^{k+1}) + \nabla f(\mathbf{x}^{k+1})$
in Eq.~\eqref{eq:error-prac} and following Theorem~\ref{thm:global-convergence}, we directly have that if $k\to\infty$, then
\begin{equation}
\mathbf{e}^{k+1}- \nabla f_{\mathbf{x}^k}^{\mu^k}(\mathbf{x}^{k+1})\in\partial r(\mathbf{x}^{k+1}).
\end{equation}
Therefore, we actually obtain the following implementable error calculation formulation for $\mathcal{E}^k$
\begin{equation}
\begin{array}{l}
\mathcal{E}^k(\mathbf{x}^{k+1}):=\mathbf{e}^{k+1}\\
=(\mu^k-1)(\mathbf{x}^{k+1}-\mathbf{v}^{k+1})-\nabla f(\mathbf{v}^{k+1}) + \nabla f(\mathbf{x}^{k+1}).
\end{array}
\end{equation}

%
%
%

%
%

\subsection{Discussions}

Intuitively, one may argue that building a deeper network should definitely result good performance. But unfortunately, many empirical evidences \cite{simonyan2014very} have suggested that the improvement cannot be trivially gained by simply adding more layers, or worse,
deeper networks even suffer from a decline on performance in some applications \cite{shen2016relay}. Therefore, it is particularly worthy of investigating the intrinsic propagation behaviors for networks with different topological structures and architectures from more solid theoretical perspective.

Indeed, our above theories have built intrinsic theoretical connections between unrolled deep models and original numerical schemes. We also investigate conditions for incorporating heuristic architectures into the proposed deep model. Therefore, the studies in this paper should provide a new perspective and introduce several powerful tools from optimization area to address the challenging but fundamental issues discussed in above paragraph.

\section{Experiments}

To verify our theoretical results and demonstrate the effectiveness of our deep models in application fields,
we apply PADNet on two real-world applications, i.e., non-blind deconvolution and single image haze removal.  All experiments are conduced on a PC with Intel Core i7 CPU at 3.4 GHz, 32 GB RAM and a NVIDIA GeForce GTX 1050 Ti GPU.


%
%

\subsection{Non-blind Deconvolution}
\begin{table*}[htb]
	\centering
	\caption{Averaged convergence results of ADMM, HQS, IPADNet and EPADNet on Eq.~\eqref{eq:deconv-gradient}.}\label{tab:converge}
	\begin{tabular}{c|cccc|cccc}
		\toprule
		Ave.&  \multicolumn{4}{c|}{Number of Iterations (denoted as $K$) } &  \multicolumn{4}{c}{$\|\mathbf{g}^K-\mathbf{g}^{\texttt{gt}}\|/\|\mathbf{g}^{\texttt{gt}}\|$} \\\midrule
		Alg. & ADMM & HQS & IPADNet & EPADNet  & ADMM & HQS  & IPADNet  & EPADNet\\ \midrule
		$\ell_1$	&29   & 33 & 6 & \textbf{5}       & 1.3146 & 1.3354 & \textbf{0.4291}& 0.4516\\
		$\ell_{0.8}$   &28     & 32 & 6 & \textbf{5}       & 1.2069 & 1.2215 & 0.4291& \textbf{0.4140}\\
		$\ell_0$   &40     & 54  & \textbf{6} &\textbf{6}       & 1.7114 & 1.7598 & \textbf{0.4291}&0.6005\\
		\bottomrule
	\end{tabular}
\end{table*}
We first consider non-blind deconvolution, which is an important task in learning and vision areas.
Specifically, given an observation $\mathbf{y}$ (e.g., image), the latent signal $\mathbf{x}$ can be processed
in a filtered domain as follows~\cite{krishnan2009fast,schmidt2014shrinkage}
$\mathcal{D}(\mathbf{y}) = \mathcal{D}(\mathbf{x})\otimes\mathbf{k} + \bm{\varepsilon}$,
where $\mathcal{D}$
is a set of filters (e.g., horizontal and vertical gradient operations), $\otimes$ denotes convolution, $\mathbf{k}$ is a point spread function and $\bm{\varepsilon}$ denotes errors/noises.
This problem can be formulated as the maximum-a-posteriori estimation
\begin{equation}
\mathbf{x}^*=\arg\max\limits_{\mathbf{x}} p(\mathbf{x}|\mathbf{y})=\arg\max\limits_{\mathbf{x}}\log p(\mathbf{y}|\mathbf{x})+\log p(\mathbf{x}).\label{eq:deconv}
\end{equation}
Here we follow typical choices to
consider $\ell_2$-fidelity (i.e., $p(\mathbf{y}|\mathbf{x}) \propto \exp(-\frac{1}{2}\|\mathcal{D}(\mathbf{x})\otimes\mathbf{k}-\mathcal{D}(\mathbf{y})\|^2)$)
and $\ell_p$-regularization (i.e., $p(\mathbf{x}) \propto \exp(-\lambda\|\mathcal{D}(\mathbf{x})\|_p^p)$, $0\leq p\leq 1$),
where $\lambda$ is the parameter. We adopt results in~\cite{zuo2013generalized} to calculate the proximal operation of general $\ell_p$-minimization. In following deconvolution experiments, we always use the set of $400$ images of size $180\times 180$ built in \cite{chen2016trainable} as our training data. Two commonly used image deblurring benchmarks respectively collected by Levin~\emph{et. al.}~\cite{levin2009understanding} (32 blurry images of size $255\times 255$) and Sun~\emph{et. al.}~\cite{patchdeblur_iccp2013}
(640 blurry images with $1\%$ Gaussian noises, sizes range from $620\times1024$ to $928\times1024$)
are used for testing.



\subsubsection{Convergence Behaviors on Gradient Domain}
The gradient of images plays very important role
in image structure analysis. Here we
first consider deconvolution on gradient domain to verify
the convergence behaviors of our designed deep models to a given energy with a simple prior. Specifically, the energy in gradient domain is defined as
\begin{equation}
\min\limits_{\mathbf{g}} \frac{1}{2}\|\mathbf{g}\otimes\mathbf{k}-\mathcal{D}(\mathbf{y})\|^2 + \lambda\|\mathbf{g}\|_p^p,\label{eq:deconv-gradient}
\end{equation}
where $\mathbf{g}$ denotes the gradient of the latent image. We first
build the basic architecture $\mathcal{G}$ as cascade of two convolutions with one RBF nonlinearity~\cite{schmidt2014shrinkage}  between them. Then we perform Alg.~\ref{alg:padm-train} based on Eq.~\eqref{eq:deconv-gradient} with $p = 0, 0.8, 1$ to respectively design three EPADNet models. We also establish an IPADNet model from  Alg.~\ref{alg:padm-train} with only the fidelity $f=\frac{1}{2}\|\mathbf{g}\otimes\mathbf{k}-\mathcal{D}(\mathbf{y})\|^2$.
To compare iteration behaviors with conventional optimization strategies, we also perform popular ADMM and Half-Quadratic Splitting (HQS)~\cite{zuo2013generalized} algorithms on Eq.~\eqref{eq:deconv-gradient} with the same $\ell_p$-regularizer
and parameters.

The averaged convergence results of compared algorithms on Levin~\emph{et. al.}' benchmark are reported in Tab.~\ref{tab:converge}. As IPDANet does not depend on $\ell_p$ functions, we just repeated its results for three cases (i.e.,  $p = 0, 0.8, 1$) in this table.

It can be seen that our designed deep models (i.e., one IPADNet and three EPADNets) need extremely less iterations but obtain more accurate estimations than conventional optimization schemes. Moreover, the performance of IPDANet is better than EPADNets regularized by $\ell_0$ and $\ell_1$, but a little worse than the $\ell_{0.8}$ energy. These results make sense because the prior learned from training data should perform better than the relatively improper handcrafted priors (e.g., $\ell_0$ and $\ell_1$ norms in this task). If the prior function can fit the data distribution well (e.g., $\ell_{0.8}$ norm here), the critical-point convergence guarantee of EPADNet will definitely result better performance, compared with the relatively weak fixed-point convergence of IPADNet.

\begin{figure*}[htb]
\centering
\includegraphics[width=0.95\textwidth]{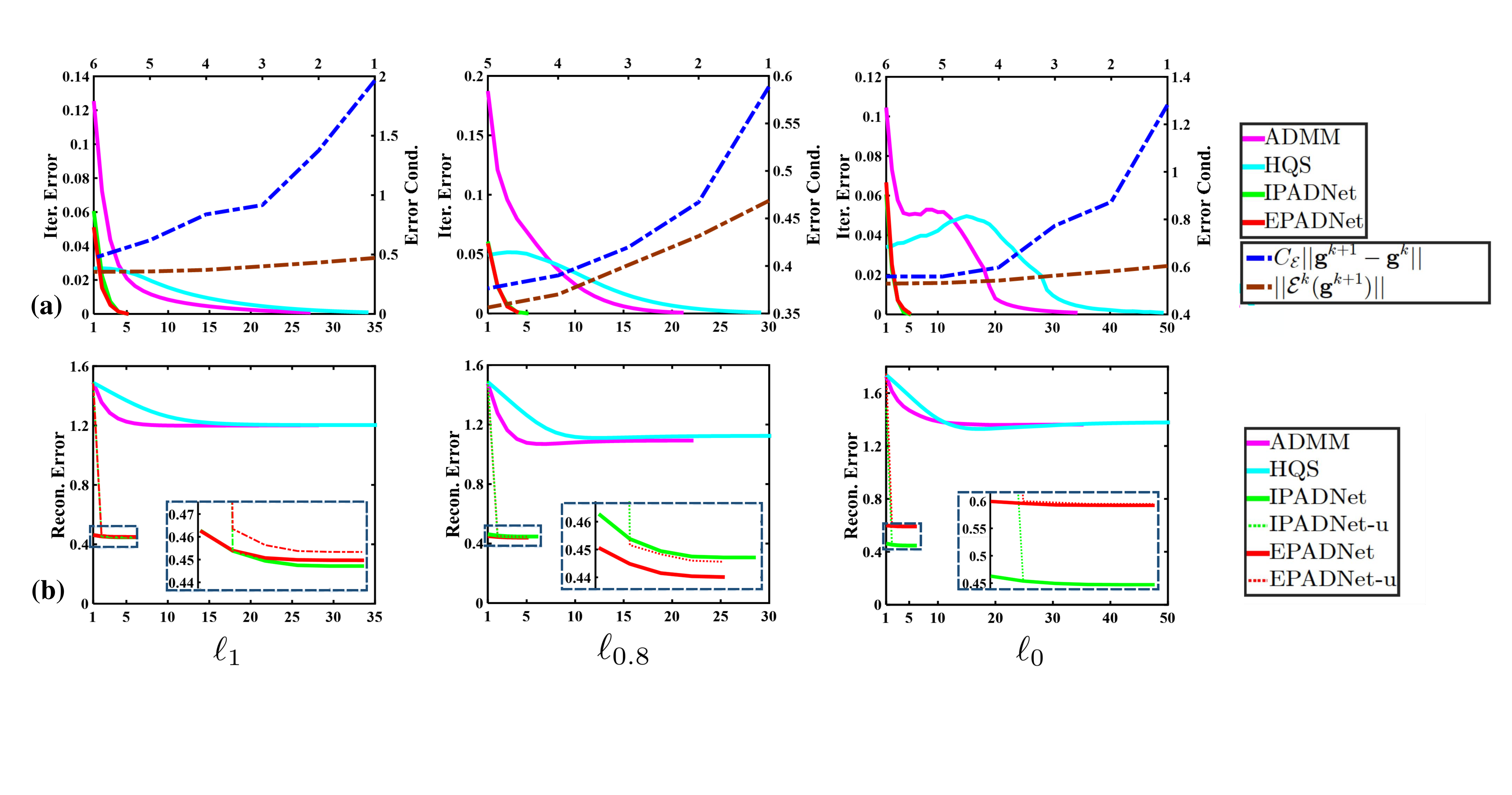}
	\caption{Convergence curves of ADMM, HQS, IPADNet and EPADNet on Eq.~\eqref{eq:deconv-gradient}. (a) Iteration Error (``Iter. Error'').  (b) Reconstruction Error (``Recon. Error''). Error Condition (``Error Cond.'') referring to EPADNet is  illustrated by dashdot curves with right and top axises on subfigures in (a). ``Recon. Error'' of the auxiliary variable (i.e., $\mathbf{u}$) of our deep models are plotted by relatively thin dotted curves on subfigures in (b). All horizontal axises denote the number of iterations.}\label{fig:converge}
\end{figure*}

\begin{table*}[htb]
	\centering
	\caption{Averaged non-blind deconvolution results on Levin~\emph{et. al.}' and Sun~\emph{et. al.}' benchmarks.}\label{tab:deconv-benchmark}
	\begin{tabular}{c|c|ccc|cccccccc}
		\toprule
		& Alg.	& TV & HL & Ours (E) & EPLL &  IDD-BM3D & MLP & CSF   & Ours (I)\\\midrule
		\multirow{2}{*}{Levin} & PSNR    & 29.38  & 30.12 & \textbf{33.41} &31.65& 31.53&31.32&32.74   & 33.37\\
		&SSIM    &0.88   &0.90  & \textbf{0.95} &0.93&0.90&0.90&0.93   &\textbf{0.95}\\\midrule
		\multirow{2}{*}{Sun}   & PSNR   &30.67  &31.03 & 32.69 & 32.44& 30.79&31.47&31.55 &\textbf{32.71}\\
		&SSIM    &0.85   &0.85  & \textbf{0.89} & 0.88&0.87&0.86&0.87   &\textbf{0.89}\\
		\bottomrule
	\end{tabular}
\end{table*}


\begin{table*}[htb]
	\centering
	\caption{Averaged single image haze removal results on Fattal's benchmark.}\label{tab:dehaze-benchmark}
	\begin{tabular}{c|ccccc|ccc}
		\toprule
		Alg. & He~\emph{et. al.} & Meng~\emph{et. al.} & Chen~\emph{et. al.} & Berman~\emph{et. al.} & Li~\emph{et. al.}&Ren~\emph{et. al.} & Cai~\emph{et. al.}   & Ours \\ \midrule
		PSNR&27.11 &26.13 &26.47 &26.09  &25.54 &24.40 &21.63 &\textbf{28.47}  \\
		SSIM &0.96  &0.95  &0.93   &0.95   &0.94 &0.94  &0.89&\textbf{0.96}  \\\midrule
		Time (s) &17.20 &6.95  &272.07  &3.73   &62.67 &5.87  &5.77  &\textbf{2.74}\\
		\bottomrule
	\end{tabular}
\end{table*}

We also plot curves of relative errors (i.e., iteration error $\|\mathbf{g}^{k+1}-\mathbf{g}^k\|/\|\mathbf{g}^k\|$ and reconstruction error $\|\mathbf{g}^k-\mathbf{g}^{\texttt{gt}}\|/\|\mathbf{g}^{\texttt{gt}}\|$) and error condition (referring to $\|\mathcal{E}^k(\mathbf{g}^{k+1})\|$ and $C_{\mathcal{E}}\|\mathbf{g}^{k+1}-\mathbf{g}^k\|$) on an example image from this benchmark in Fig.~\ref{fig:converge}, where $\mathbf{g}^{\texttt{gt}}$ denotes the ground-truth image gradient.
To provide more readable illustrations of convergence behaviors, here all relative errors are plotted starting from $k=1$. We also show zoomed in curve comparisons of our two deep models in Fig.~\ref{fig:converge} (b). Notice that we indeed only have one implicit
deep model for this task. But to compare its performance with methods based on different
$\ell_p$ energies, we just repeatedly plot its relative errors (as green curves) in multiple subfigures.

It is observed in Fig.~\ref{fig:converge} (a) that our deep models always converged within $5$-$6$ iterations, while
both ADMM and HQS needed dozens of steps to stop their iterations. The dashdot curves in Fig.~\ref{fig:converge} (a) show that the designed EPADNet satisfied the constraint of errors in Condition~\ref{con:error} all the time, thus the global convergence to the critical-point of Eq.~\eqref{eq:deconv-gradient} can be experimentally guaranteed. All these results verified our proved theories.
We can further see in Fig.~\ref{fig:converge} (b) that propagations of our two deep models (solid red and green curves) had obtained significantly lower reconstruction errors than conventional algorithms even just after the first iteration (i.e., the initial points of these curves). This is because built-in networks actually learned a direct descent direction toward the desired solutions, which demonstrated the superiority of our framework again.

\subsubsection{Explicit / Implicit PADNet on Image Domain}
Non-blind deconvolution on image domain is commonly formulated as the following energy minimization task~\cite{li2013efficient,krishnan2009fast,schmidt2014shrinkage}
\begin{equation}
\min\limits_{\mathbf{x}, \mathbf{g}}\frac{1}{2}f_{\texttt{ex}}(\mathbf{x}, \mathbf{g};\mathbf{y}) + \lambda\|\mathbf{g}\|_p^p,\label{eq:deconv-image}
\end{equation}
in which the fidelity $f_{\texttt{ex}}$ can be formulated as
\begin{equation}
f_{\texttt{ex}}(\mathbf{x}, \mathbf{g};\mathbf{y}):=\inf\limits_{\mathbf{x}}\left\{\|\mathbf{x}\otimes\mathbf{k}-\mathbf{y}\|^2 + \beta\|\mathcal{D}(\mathbf{x})-\mathbf{g}\|^2\right\}.
\end{equation}
Here $\beta$ is a penalty parameter, $\mathbf{x}$ and $\mathbf{g}$ are variables in image and gradient domains, respectively.

In this part, we build an explicit PADNet using Eq.~\eqref{eq:deconv-image} to pursuit $\mathbf{x}$, in which we set $p=0.8$ and introduce an additional linear layer derived by $\nabla_{\mathbf{x}} f_{\texttt{ex}} = 0$ to transfer variables from gradient domain to image domain.
In contrast, by simply defining $f_{\texttt{im}}(\mathbf{x};\mathbf{y})=\frac{1}{2}\|\mathbf{x}\otimes\mathbf{k}-\mathbf{y}\|^2$ and discarding explicit $\ell_p$-priors,
we can also design an implicit PADNet to learn priors from training data for this task.
Here the basic architecture $\mathcal{G}$ (used in our deep models) consists of $7$ convolution layers. The ReLU nonlinearities are added between each two linear layers accordingly and batch normalizations (BN)~\cite{ioffe2015batch} are also introduced for convolution operations from $2$-nd to $6$-th linear layers.

We compare performances of our two deep models against state-of-the-art algorithms, including TV~\cite{li2013efficient}, HL~\cite{krishnan2009fast}, EPLL~\cite{zoran2011learning}, IDD-BM3D~\cite{danielyan2012bm3d}, MLP~\cite{schuler2013machine} and CSF~\cite{schmidt2014shrinkage} on both standard Levin~\emph{et. al.}' and  more challenging Sun~\emph{et. al.}' benchmarks. The averaged quantitative results (i.e., PSNR and SSIM), are reported in Tab.~\ref{tab:deconv-benchmark}, in which ``(E)'' and ``(I)''  denote algorithms based on explicit and implicit PADNet, respectively. We can recognize that ``Ours (E)'' and the works in \cite{li2013efficient,krishnan2009fast} actually all address this task by optimizing Eq.~\eqref{eq:deconv-image}.
Thanks to built-in networks, we achieved much better performance than conventional optimization approaches. We further observed that discriminative learning approaches \cite{schmidt2014shrinkage,schuler2013machine} also performed well as they learn adaptive networks from training data. Overall, the results of our two algorithms are better than other compared approaches. The PSNR score of ``Ours (E)''
is even higher than that of ``Ours (I)'' on standard Levin \emph{et. al.}'s dataset. We argue that this is reasonable because $\ell_{0.8}$ prior actually has been powerful enough for relatively simple test images. While ``Ours (I)'' obtained the best quantitative results on Sun \emph{et. al.}'s dataset, which demonstrated that our prior-and-data aggregated framework is especially more efficient on real-world challenging applications (see Remark~\ref{remark:fixed-point}).

\subsection{Single Image Haze Removal}
Finally, we evaluate PADNet on the task of single image haze removal, which is a challenging real-world vision
application. Most existing works address this task as estimating the latent scene radiance $\mathbf{J}$
from given hazy observation $\mathbf{I}$ from the following linear interpolation formula
\begin{equation}
\mathbf{I}(x)=t(x)\mathbf{J}(x) + (1-t(x))\mathbf{A},\label{eq:haze}
\end{equation}
where $t$ is transmission, $\mathbf{A}$ is global atmospheric light and $x$ denotes the pixel index.

It is known that transmission $t$ expresses the relative portion
of light that managed to survive the entire path between the
observer and a surface point in the scene without being
scattered~\cite{fattal2014dehazing}. With Eq.~\eqref{eq:haze},  we have that estimating accurate transmission
map $t$ plays the core role in this task. However,
due to multiple solutions exist for a single hazy image, the
problem is highly ill-posed.
Recent works often design their models based on different perspectives on transmissions within the following prior regularized energy
\begin{equation}
\begin{array}{l}
\min\limits_{\mathbf{t}} \frac{1}{2}\|\mathbf{t}-\tilde{\mathbf{t}}\|^2 + \lambda r(\mathbf{t}), \\
 \mbox{e.g.,}
\left\{
\begin{array}{l}
r_{\mathtt{TGV}}(\mathbf{t})=\|\nabla \mathbf{t}-\mathbf{z}\|_1+\beta\|\nabla \mathbf{z}\|_{1},
\\ \mbox{\cite{chen2016robust}},
\\
r_{\mathtt{MRF}}(\mathbf{t})=\sum_{j}\|\mathbf{w}_j\odot\left(\mathbf{d}_j\otimes\mathbf{t}\right)\|_1, \\ \mbox{\cite{meng2013efficient}},
\end{array}
\right.
\end{array}
\end{equation}
where $\mathbf{t}$ and $\tilde{\mathbf{t}}$ respectively denote the discrete transmission vector
and its propagation guidance. The regularization $r$ can be derived based on different tools, e.g., Total Generalized Variation (TGV)~\cite{chen2016robust} and Markov Random Field (MRF)~\cite{meng2013efficient}. In $r_{\texttt{MRF}}$, $\mathbf{z}$ denotes an auxiliary variable, $\odot$ is Hadamard product and $\{\mathbf{w}_j\}$ are weight vectors for local filters $\{\mathbf{d}_j\}$.

In this part, we first utilize implicit strategy to design PADNet based on fidelity $f(\mathbf{t})=\frac{1}{2}\|\mathbf{t}-\tilde{\mathbf{t}}\|^2$ (with the same guidance $\tilde{\mathbf{t}}$ defined in \cite{meng2013efficient}) to estimate transmission
and then recover the latent scene radiance from Eq.~\eqref{eq:haze} as that in \cite{he2011single,chen2016robust,meng2013efficient}. We build basic architecture $\mathcal{G}$ with $17$ convolution
layers (ReLU and BN operations are incorporated using the same strategy as that in above image deconvolution task) and train it on synthetic hazy images~\cite{ren2016single} for our deep model.


We evaluate the performance of our deep model together with five existing handcrafted-prior based algorithms  (i.e.,~\cite{he2011single,meng2013efficient,chen2016robust,berman2016non,li2014contrast}) and two empirically designed deep networks (i.e., ~\cite{ren2016single,cai2016dehazenet})\footnote{In this subsection, we always denote these methods as He~\emph{et. al.}, Meng~\emph{et. al.}, Chen~\emph{et. al.}, Berman~\emph{et. al.}, Li~\emph{et. al.}, Ren~\emph{et. al.} and Cai~\emph{et. al.}, respectively.}  on the commonly used Fattal's benchmark~\cite{fattal2008single}, which consists of $11$ challenging hazy images, including architecture, natural scenery and indoor scene.
The averaged quantitative results, including PSNR, SSIM and running time in seconds (denoted as ``Time (s)''), are given in Tab.~\ref{tab:dehaze-benchmark}.
Two empirically designed networks in~\cite{cai2016dehazenet,ren2016single} performed better than most conventional prior-based methods. Though obtained good dehazing results, the work in \cite{chen2016robust} has the longest running time. Our proposed deep model achieved the best performance among all compared algorithms on this benchmark. This is mainly because that PADNet can successfully fuse cues from both human perspectives and training data to estimate haze distributions. Furthermore, the speed of PADNet is the fastest among all compared methods, which also verified the efficiency of our framework.

\section{Conclusions}

This paper proposed a novel framework, named proximal alternating direction network (PADNet), to design
deep models for different learning tasks. Our theoretical results first showed that we can utilize empirically designed architectures
to build globally converged PADNet for the given energy minimization model. We further proved that a converged PADNet can also be designed by learning priors from training data. At last we experimentally verified our analysis and demonstrated promising results of PADNet on different real-world applications.

\section{Acknowledgments}

This work is partially supported by the National Natural Science Foundation of China (Nos.  61672125, 61632019, 61432003, 61572096 and 61733002), and the Hong Kong Scholar Program (No. XJ2015008). Dr. Liu is also a visiting researcher with Shenzhen Key Laboratory of Media Security, Shenzhen University, Shenzhen 518060

\section{Supplemental Material}

\subsection{Preliminaries}\label{sec:pre}
The following definitions and lemmas have been widely known in variational analysis and optimization. More details can also be found in \cite{rockafellar2009variational,attouch2010proximal} and references therein.
\begin{defi} Here we give necessary definitions, including proper and lower semi-continuous, coercive and  semi-algebraic.
	\begin{itemize}
		\item A function $r: \mathbb{R}^n\to(-\infty, +\infty]$ is said to be proper and lower semi-continuous if $\texttt{dom}(r)\neq \emptyset$, where $\texttt{dom}(r) :=\{\mathbf{x}\in\mathbb{R}^n: r(\mathbf{x})<+\infty\}$ and $\liminf_{\mathbf{x}\to\mathbf{y}}r(\mathbf{x})\geq r(\mathbf{y})$ at any point $\mathbf{y}\in\texttt{dom}(r)$.
		\item A function $\mathcal{F}$ is said to be coercive, if $\mathcal{F}$ is bounded from below and $\mathcal{F}\to\infty$ if $\|\mathbf{x}\|\to\infty$, where $\|\cdot\|$ is the $\ell_2$ norm.
		\item A subset $S$ of $\mathbb{R}^n$ is a real semi-algebraic set if there exist a finit number of real polynomial functions
		$g_{ij}, h_{ij}: \mathbb{R}^n\to\mathbb{R}$ such that
		\begin{equation}
		S = \bigcup\limits_{j=1}^p\bigcap\limits_{i=1}^q\left\{\mathbf{x}\in\mathbb{R}^n: g_{ij}(\mathbf{x})=0 \ \mbox{and} \ h_{ij}(\mathbf{x})<0\right\}.
		\end{equation}
		A function $r:\mathbb{R}^n\to(-\infty,\infty]$ is called semi-algebraic if its graph $\{(\mathbf{x},z)\in\mathbb{R}^{n+1}:
		r(\mathbf{x})=z\}$ is a semi-algebraic subset of $\mathbb{R}^{n+1}$.
	\end{itemize}
\end{defi}
Indeed, many functions arising in learning areas, including $\ell_0$ norm, rational $\ell_p$
norms (i.e., $p=p_1/p_2$ with positive integers $p_1$ and $p_2$)
and their finite sums or products, are all semi-algebraic.

\begin{lemma}\label{lem:lim-sub-diff} Here we list some necessary properties will be used in our following proofs.
	\begin{itemize}
		\item In the nonsmooth context, the Fermat's rule remains unchanged. That is, if $\mathbf{x}\in\mathbb{R}^n$ is
		a local minimizer of $\mathcal{F}$, then $0\in\partial \mathcal{F}(\mathbf{x})$.
		\item If $f$ is a continuously differentiable function, then $\partial(f+r)=\nabla f + \partial r$.
		\item Let $r$ be proper and lower semi-continuous and $\{(\mathbf{x}^k,\mathbf{g}_{\mathbf{x}}^k)\}$ be a sequence
		in the graph of $\partial r$. If $(\mathbf{x}^k,\mathbf{g}_{\mathbf{x}}^k)\to(\mathbf{x}^*,\mathbf{g}_{\mathbf{x}}^*)$ and $r(\mathbf{x}^k)\to r(\mathbf{x}^*)$ as $k\to\infty$, then $(\mathbf{x}^*,\mathbf{g}_{\mathbf{x}}^*)$ is also in the graph of $\partial r$.
	\end{itemize}	
\end{lemma}

Finally, we recall that a critical point of a function $\mathcal{F}$ is a point $\mathbf{x}$ in the domain of $\mathcal{F}$,
whose subdifferential $\partial\mathcal{F}(\mathbf{x})$ contains $0$.

\subsection{Proof of Theorem~\ref{thm:global-convergence}}\label{sec:thm1}

\begin{proof}
	The error condition of $\mathbf{x}^{k+1}$ in \eqref{eq:error} can be equivalently reformulated as
	\begin{equation}
	\mathbf{x}^{k+1}\in\arg\min\limits_{\mathbf{x}} f_{\mathbf{x}^k}^{\mu^k}(\mathbf{x}) + r(\mathbf{x}) - \left\langle\mathbf{x},\mathbf{e}^{k+1}\right\rangle,\label{eq:reformulate-error}
	\end{equation}
	where $\mathbf{e}^{k+1}:=\mathcal{E}^k(\mathbf{x}^{k+1})$.
	Therefore, the propagation behavior of Alg.~\ref{alg:padm-train} actually should be understood as exactly solving
	Eq.~\eqref{eq:reformulate-error} to output $\mathbf{x}^{k+1}$ at $k$-th iteration. Thus
	\begin{equation}
	\begin{array}{l}
	\quad\mathcal{F}(\mathbf{x}^{k}) - \langle\mathbf{x}^{k},\mathbf{e}^{k+1}\rangle \\
	\geq \mathcal{F}(\mathbf{x}^{k+1}) - \langle\mathbf{x}^{k+1},\mathbf{e}^{k+1}\rangle + \frac{\mu^k}{2}\|\mathbf{x}^{k+1}-\mathbf{x}^{k}\|^2\\
	\Rightarrow \mathcal{F}(\mathbf{x}^{k})-\mathcal{F}(\mathbf{x}^{k+1}) \\
	\geq \frac{\mu^k}{2}\|\mathbf{x}^{k+1}-\mathbf{x}^{k}\|^2 - \langle \mathbf{x}^{k+1}-\mathbf{x}^{k}, \mathbf{e}^{k+1}\rangle\\
	\geq \frac{\mu^k}{2}\|\mathbf{x}^{k+1}-\mathbf{x}^{k}\|^2 - \frac{1}{2\rho}\|\mathbf{x}^{k+1}-\mathbf{x}^{k}\|^2- \frac{\rho}{2}\|\mathbf{e}^{k+1}\|^2\\
	\geq \left(\frac{\mu^k}{2}-\frac{\eta C_{\mathcal{E}}^2}{2}-\frac{1}{2\eta}\right)\|\mathbf{x}^{k+1}-\mathbf{x}^{k}\|^2 \\	=\left(\frac{\mu^k}{4}-\frac{C_{\mathcal{E}}^2}{\mu^k}\right)\|\mathbf{x}^{k+1}-\mathbf{x}^{k}\|^2,\label{eq:descent}
	\end{array}
	\end{equation}
	in which $\eta$ can be any positive constant but the last equality only holds with $\eta = 2/\mu^k$.
	So we have $\mathcal{F}(\mathbf{x}^{k+1})\leq\mathcal{F}(\mathbf{x}^0)<\infty$.
	As $\mathcal{F}$ is coercive, we also have that $\{\mathbf{x}^k\}$ is bounded.
	Furthermore, as $\{\mathcal{F}(\mathbf{x}^k)\}$ is nonincreasing, it converges to a constant (denoted as $\mathcal{F}^*$). So summing \eqref{eq:descent} from $0$ to $\infty$ leads to
	\begin{equation} \min\limits_k\left\{\frac{\mu^k}{4}-\frac{C_{\mathcal{E}}^2}{\mu^k}\right\}\sum\limits_{k=0}^{\infty}\|\mathbf{x}^{k+1}-\mathbf{x}^{k}\|^2\leq \mathcal{F}(\mathbf{x}^0)-\mathcal{F}^*<\infty.
	\end{equation}
	Using $2C_{\mathcal{E}} < \mu^k <\infty$, we have $\|\mathbf{x}^{k+1}-\mathbf{x}^k\|\to 0$
	if $k\to\infty$.
	
	Following \eqref{eq:error}, we have
	\begin{equation}
	\mathbf{e}^{k+1} - \mu^k(\mathbf{x}^{k+1}-\mathbf{x}^k)\in\partial \mathcal{F}(\mathbf{x}^{k+1}).
	\end{equation}
	Then it is straightforward that
	\begin{equation}
	\begin{array}{l}
	\quad\|\partial \mathcal{F}(\mathbf{x}^{k+1})\| \\
	=\|\mathbf{e}^{k+1} - \mu^k(\mathbf{x}^{k+1}-\mathbf{x}^k)\| \\
	\leq (\mu^k + C_{\mathcal{E}})\|\mathbf{x}^{k+1}-\mathbf{x}^k\|\to 0, \ \mbox{if} \ k \to \infty.
	\end{array}\label{eq:gradient}
	\end{equation}
	
	Let $\mathbf{x}^*$ be any cluster point of $\{\mathbf{x}^k\}$, i.e., $\mathbf{x}^{k_j}\to\mathbf{x}^*$ if $j\to\infty$.
	Since $r$ is lower semi-continuous, we  have
	\begin{equation}
	\liminf\limits_{j\to\infty} r(\mathbf{x}^{k_j}) \geq r(\mathbf{x}^*).
	\end{equation}
	From \eqref{eq:reformulate-error}, we have
	\begin{equation}
	\begin{array}{l}
	\quad f_{\mathbf{x}^k}^{\mu^k}(\mathbf{x}^{k+1}) + r(\mathbf{x}^{k+1}) - \left\langle\mathbf{x}^{k+1},\mathbf{e}^{k+1}\right\rangle \\
	\leq f_{\mathbf{x}^k}^{\mu^k}(\mathbf{x}^*) + r(\mathbf{x}^*) - \left\langle\mathbf{x}^*,\mathbf{e}^{k+1}\right\rangle\\
	\Rightarrow
	r(\mathbf{x}^{k+1}) \\
	\leq r(\mathbf{x}^*) + f_{\mathbf{x}^k}^{\mu^k}(\mathbf{x}^*) - f_{\mathbf{x}^k}^{\mu^k}(\mathbf{x}^{k+1})  +\left\langle\mathbf{x}^{k+1}-\mathbf{x}^{*},\mathbf{e}^{k+1}\right\rangle\\
	\leq r(\mathbf{x}^*) + f_{\mathbf{x}^k}^{\mu^k}(\mathbf{x}^*) - f_{\mathbf{x}^k}^{\mu^k}(\mathbf{x}^{k+1})  \\  \quad +C_{\mathcal{E}}\|\mathbf{x}^{k+1}-\mathbf{x}^{*}\|\cdot\|\mathbf{x}^{k+1}-\mathbf{x}^{k}\|.
	\end{array}\label{eq:continuity-r}
	\end{equation}
	Setting $k=k_j-1$ in \eqref{eq:continuity-r} and letting $j\to\infty$, we obtain
	\begin{equation}
	\limsup\limits_{j\to\infty}r(\mathbf{x}^{k_j})
	\leq r(\mathbf{x}^*),
	\end{equation}
	in which we used the facts that $f^k$ is continuous and $\{\mathbf{x}^k\}$ bounded. So we have
	that $r(\mathbf{x}^{k_j})$ tends to $r(\mathbf{x}^*)$ as $j\to\infty$. This together with the continuity of $f$ directly results
	to
	\begin{equation}
	\lim\limits_{j\to\infty}\mathcal{F}(\mathbf{x}^{k_j}) = \mathcal{F}(\mathbf{x}^*).\label{eq:continuity-F}
	\end{equation}
	Using \eqref{eq:gradient}, \eqref{eq:continuity-F} and Lemma~\ref{lem:lim-sub-diff}, we have that $0\in\partial\mathcal{F}(\mathbf{x}^*)$. So any cluster point of $\{\mathbf{x}^k\}$ is a critical point of Eq.~\eqref{eq:model}.
	
	If we further assume that $\mathcal{F}$ is a semi-algebraic function, then
	it satisfies the well-known Kurdyka-{\L}ojasiewicz property~\cite{rockafellar2009variational,attouch2010proximal}. So we have that $\sum_{k=0}^{\infty}\|\mathbf{x}^k-\mathbf{x}^{k-1}\|<\infty$ following \cite{attouch2010proximal}. This implies that $\{\mathbf{x}^k\}$ is a Cauchy sequence
	and hence a convergent sequence. Since we have proved in above that $\mathbf{x}^{k_j}\to\mathbf{x}^*$ if $j\to\infty$, we finally have
	$\mathbf{x}^{k}\to\mathbf{x}^*$ if $k\to\infty$, which completes the proof.
\end{proof}

\subsection{Proof of Theorem~\ref{thm:fixed-point-convergence}}\label{sec:thm2}

\begin{proof}
	The updating scheme of $\mathbf{u}$ in Alg.~\ref{alg:padm-train} implies
	\begin{equation}
	\mathbf{u} - (\mathbf{v}^{k}-\bm{\lambda}^k) = -\frac{1}{\rho^k}\nabla f_{\mathbf{x}^k}^{\mu^k}(\mathbf{u}).
	\end{equation}
	Moreover, it is easy to check that $\nabla f_{\mathbf{x}^k}^{\mu^k}$ is bounded, i.e. there exists $M<\infty$ such that $\|\nabla f_{\mathbf{x}^k}^{\mu^k}(\mathbf{u})\|\leq M$.
	So at $k$-th iteration, we have
	\begin{equation}
	\|\mathbf{u}^{k+1} - (\mathbf{v}^{k}-\bm{\lambda}^k)\|=\frac{1}{\rho^k}\|\nabla f_{\mathbf{x}^k}^{\mu^k}(\mathbf{u})\| \leq \frac{M}{\rho^k}.\label{eq:x-u-bound}
	\end{equation}
	Next, define $\tilde{\mathbf{v}}^k=\mathbf{u}^{k+1}+\bm{\lambda}^k$. Then following Steps \ref{step:v-net} and \ref{setp:v-updata} in Alg.~\ref{alg:padm-train}
	and the bounded constraint on $\mathcal{N}_{\alpha}$, we have
	\begin{equation}
	\begin{array}{l}
	\quad\|\mathbf{v}^{k+1}-(\mathbf{u}^{k+1}+\bm{\lambda}^k)\| \\
	=\|\mathcal{N}_{\alpha^k}(\tilde{\mathbf{v}}^k)-\tilde{\mathbf{v}}^k\| \\
	\leq C_{\mathcal{N}}\alpha^k \\
	= C_{\mathcal{N}}\sqrt{\frac{1}{\rho^k}}.
	\label{eq:u-x-bound}
	\end{array}
	\end{equation}
	Using Eq.~\eqref{eq:x-u-bound} and \eqref{eq:u-x-bound}, we can show
	\begin{equation}
	\begin{array}{l}
	\quad\|\mathbf{v}^{k+1}-\mathbf{v}^k\| \\
	\leq \|\mathbf{v}^{k+1}-\tilde{\mathbf{v}}^k\| + \|\tilde{\mathbf{v}}^k-\mathbf{v}^{k}\|\\
	\leq C_{\mathcal{N}}\sqrt{\frac{1}{\rho^k}} + \frac{M}{\rho^k} \\ =\sqrt{\frac{1}{\rho^k}}\left(C_{\mathcal{N}}+M\sqrt{\frac{1}{\rho^k}}\right)\\
	\leq \left(C_{\mathcal{N}}+M\sqrt{\frac{1}{\rho^0}}\right)\sqrt{\frac{1}{\rho^k}}.
	\end{array}
	\end{equation}
	As for $\bm{\lambda}^{k+1}$, we have
	\begin{equation}
	\begin{array}{l}
	\quad\|\bm{\lambda}^{k+1}\|=\|\bm{\lambda}^{k}+(\mathbf{u}^{k+1}-\mathbf{v}^{k+1})\|\\
	=\|\bm{\lambda}^{k}+\mathbf{u}^{k+1}- \tilde{\mathbf{v}}^k-(\mathcal{N}_{\alpha^k}(\tilde{\mathbf{v}}^k) - \tilde{\mathbf{v}}^k) \| \\
	=\|\mathcal{N}_{\alpha^k}(\tilde{\mathbf{v}}^k) - \tilde{\mathbf{v}}^k\|\leq C_{\mathcal{N}}\sqrt{\frac{1}{\rho^k}}\\
	\Rightarrow \|\bm{\lambda}^{k+1}-\bm{\lambda}^{k}\| \leq \|\bm{\lambda}^{k+1}\| + \|\bm{\lambda}^{k}\|\leq 2C_{\mathcal{N}}\sqrt{\frac{1}{\rho^k}},
	\end{array}
	\end{equation}
	Following Step \ref{step:lambda} in Alg.~\ref{alg:padm-train}, we have
	\begin{equation}
	\begin{array}{l}
	\quad\|\mathbf{u}^{k+1}-\mathbf{u}^{k}\| \\
	=\|(\bm{\lambda}^{k+1}-\bm{\lambda}^{k} + \mathbf{v}^{k+1})-(\bm{\lambda}^{k}-\bm{\lambda}^{k-1} + \mathbf{v}^{k})\|\\
	\leq \|\bm{\lambda}^{k+1}-\bm{\lambda}^{k}\| + \|\bm{\lambda}^{k}-\bm{\lambda}^{k-1}\| + \|\mathbf{v}^{k+1}-\mathbf{v}^{k}\|\\
	\leq 2C_{\mathcal{N}}\sqrt{\frac{1}{\rho^k}} + 2C_{\mathcal{N}}\sqrt{\frac{1}{\rho^{k-1}}} +  \sqrt{\frac{1}{\rho^k}}\left(C_{\mathcal{N}}+M\sqrt{\frac{1}{\rho^0}}\right)\\
	\leq\left((3+2\gamma)C_{\mathcal{N}}+\frac{M}{\rho^0}\right)\sqrt{\frac{1}{\rho^k}}.
	\end{array}
	\end{equation}
	
	So $\{(\mathbf{u}^k,\mathbf{v}^k,\bm{\lambda}^k)\}$ is a Cauchy sequence and
	hence there exists fixed point $(\mathbf{u}^*,\mathbf{v}^*,\bm{\lambda}^*)$ such that
	$(\mathbf{u}^k,\mathbf{v}^k,\bm{\lambda}^k)\to(\mathbf{u}^*,\mathbf{v}^*,\bm{\lambda}^*)$ if $k\to\infty$, which completes the proof.
\end{proof}

\bibliographystyle{aaai}
\bibliography{egbib}

\begin{thebibliography}{}

\bibitem[\protect\citeauthoryear{Andrychowicz \bgroup et al\mbox.\egroup
  }{2016}]{andrychowicz2016learning}
Andrychowicz, M.; Denil, M.; Gomez, S.; Hoffman, M.~W.; Pfau, D.; Schaul, T.;
  and de~Freitas, N.
\newblock 2016.
\newblock Learning to learn by gradient descent by gradient descent.
\newblock In {\em NIPS},  3981--3989.

\bibitem[\protect\citeauthoryear{Attouch \bgroup et al\mbox.\egroup
  }{2010}]{attouch2010proximal}
Attouch, H.; Bolte, J.; Redont, P.; and Soubeyran, A.
\newblock 2010.
\newblock Proximal alternating minimization and projection methods for
  nonconvex problems: An approach based on the kurdyka-{\l}ojasiewicz
  inequality.
\newblock {\em Mathematics of Operations Research} 35(2):438--457.

\bibitem[\protect\citeauthoryear{Berman, Avidan, and
  others}{2016}]{berman2016non}
Berman, D.; Avidan, S.; et~al.
\newblock 2016.
\newblock Non-local image dehazing.
\newblock In {\em CVPR},  1674--1682.

\bibitem[\protect\citeauthoryear{Cai \bgroup et al\mbox.\egroup
  }{2016}]{cai2016dehazenet}
Cai, B.; Xu, X.; Jia, K.; Qing, C.; and Tao, D.
\newblock 2016.
\newblock Dehazenet: An end-to-end system for single image haze removal.
\newblock {\em IEEE TIP} 25(11):5187--5198.

\bibitem[\protect\citeauthoryear{Chen and Pock}{2017}]{chen2016trainable}
Chen, Y., and Pock, T.
\newblock 2017.
\newblock Trainable nonlinear reaction diffusion: A flexible framework for fast
  and effective image restoration.
\newblock {\em IEEE TPAMI} 39(6):1256--1272.

\bibitem[\protect\citeauthoryear{Chen, Do, and Wang}{2016}]{chen2016robust}
Chen, C.; Do, M.~N.; and Wang, J.
\newblock 2016.
\newblock Robust image and video dehazing with visual artifact suppression via
  gradient residual minimization.
\newblock In {\em ECCV},  576--591.

\bibitem[\protect\citeauthoryear{Chouzenoux, Pesquet, and
  Repetti}{2016}]{chouzenoux2016block}
Chouzenoux, E.; Pesquet, J.-C.; and Repetti, A.
\newblock 2016.
\newblock A block coordinate variable metric forward--backward algorithm.
\newblock {\em Journal of Global Optimization} 66(3):457--485.

\bibitem[\protect\citeauthoryear{Danielyan, Katkovnik, and
  Egiazarian}{2012}]{danielyan2012bm3d}
Danielyan, A.; Katkovnik, V.; and Egiazarian, K.
\newblock 2012.
\newblock Bm3d frames and variational image deblurring.
\newblock {\em IEEE TIP} 21(4):1715--1728.

\bibitem[\protect\citeauthoryear{Fattal}{2008}]{fattal2008single}
Fattal, R.
\newblock 2008.
\newblock Single image dehazing.
\newblock {\em ACM Transactions on Graphics (TOG)} 27(3):72.

\bibitem[\protect\citeauthoryear{Fattal}{2014}]{fattal2014dehazing}
Fattal, R.
\newblock 2014.
\newblock Dehazing using color-lines.
\newblock {\em ACM Transactions on Graphics (TOG)} 34(1):13.

\bibitem[\protect\citeauthoryear{Gregor and LeCun}{2010}]{gregor2010learning}
Gregor, K., and LeCun, Y.
\newblock 2010.
\newblock Learning fast approximations of sparse coding.
\newblock In {\em ICML},  399--406.

\bibitem[\protect\citeauthoryear{He \bgroup et al\mbox.\egroup
  }{2016}]{he2016deep}
He, K.; Zhang, X.; Ren, S.; and Sun, J.
\newblock 2016.
\newblock Deep residual learning for image recognition.
\newblock In {\em CVPR},  770--778.

\bibitem[\protect\citeauthoryear{He, Sun, and Tang}{2011}]{he2011single}
He, K.; Sun, J.; and Tang, X.
\newblock 2011.
\newblock Single image haze removal using dark channel prior.
\newblock {\em IEEE TPAMI} 33(12):2341--2353.

\bibitem[\protect\citeauthoryear{Ioffe and Szegedy}{2015}]{ioffe2015batch}
Ioffe, S., and Szegedy, C.
\newblock 2015.
\newblock Batch normalization: Accelerating deep network training by reducing
  internal covariate shift.
\newblock In {\em ICML},  448--456.

\bibitem[\protect\citeauthoryear{Krishnan and Fergus}{2009}]{krishnan2009fast}
Krishnan, D., and Fergus, R.
\newblock 2009.
\newblock Fast image deconvolution using hyper-laplacian priors.
\newblock In {\em NIPS},  1033--1041.

\bibitem[\protect\citeauthoryear{Krizhevsky, Sutskever, and
  Hinton}{2012}]{krizhevsky2012imagenet}
Krizhevsky, A.; Sutskever, I.; and Hinton, G.~E.
\newblock 2012.
\newblock Imagenet classification with deep convolutional neural networks.
\newblock In {\em NIPS},  1097--1105.

\bibitem[\protect\citeauthoryear{Levin \bgroup et al\mbox.\egroup
  }{2009}]{levin2009understanding}
Levin, A.; Weiss, Y.; Durand, F.; and Freeman, W.~T.
\newblock 2009.
\newblock Understanding and evaluating blind deconvolution algorithms.
\newblock In {\em CVPR},  1964--1971.

\bibitem[\protect\citeauthoryear{Li \bgroup et al\mbox.\egroup
  }{2013}]{li2013efficient}
Li, C.; Yin, W.; Jiang, H.; and Zhang, Y.
\newblock 2013.
\newblock An efficient augmented lagrangian method with applications to total
  variation minimization.
\newblock {\em Computational Optimization and Applications} 56(3):507--530.

\bibitem[\protect\citeauthoryear{Li \bgroup et al\mbox.\egroup
  }{2014}]{li2014contrast}
Li, Y.; Guo, F.; Tan, R.~T.; and Brown, M.~S.
\newblock 2014.
\newblock A contrast enhancement framework with jpeg artifacts suppression.
\newblock In {\em ECCV},  174--188.

\bibitem[\protect\citeauthoryear{Lin, Liu, and Su}{2011}]{lin2011linearized}
Lin, Z.; Liu, R.; and Su, Z.
\newblock 2011.
\newblock Linearized alternating direction method with adaptive penalty for
  low-rank representation.
\newblock In {\em NIPS},  612--620.

\bibitem[\protect\citeauthoryear{Meng \bgroup et al\mbox.\egroup
  }{2013}]{meng2013efficient}
Meng, G.; Wang, Y.; Duan, J.; Xiang, S.; and Pan, C.
\newblock 2013.
\newblock Efficient image dehazing with boundary constraint and contextual
  regularization.
\newblock In {\em ICCV},  617--624.

\bibitem[\protect\citeauthoryear{Parikh, Boyd, and
  others}{2014}]{parikh2014proximal}
Parikh, N.; Boyd, S.; et~al.
\newblock 2014.
\newblock Proximal algorithms.
\newblock {\em Foundations and Trends{\textregistered} in Optimization}
  1(3):127--239.

\bibitem[\protect\citeauthoryear{Ren \bgroup et al\mbox.\egroup
  }{2016}]{ren2016single}
Ren, W.; Liu, S.; Zhang, H.; Pan, J.; Cao, X.; and Yang, M.-H.
\newblock 2016.
\newblock Single image dehazing via multi-scale convolutional neural networks.
\newblock In {\em ECCV},  154--169.

\bibitem[\protect\citeauthoryear{Rockafellar and
  Wets}{2009}]{rockafellar2009variational}
Rockafellar, R.~T., and Wets, R. J.-B.
\newblock 2009.
\newblock {\em Variational analysis}, volume 317.
\newblock Springer Science \& Business Media.

\bibitem[\protect\citeauthoryear{Schmidt and Roth}{2014}]{schmidt2014shrinkage}
Schmidt, U., and Roth, S.
\newblock 2014.
\newblock Shrinkage fields for effective image restoration.
\newblock In {\em CVPR},  2774--2781.

\bibitem[\protect\citeauthoryear{Schuler \bgroup et al\mbox.\egroup
  }{2013}]{schuler2013machine}
Schuler, C.~J.; Christopher~Burger, H.; Harmeling, S.; and Scholkopf, B.
\newblock 2013.
\newblock A machine learning approach for non-blind image deconvolution.
\newblock In {\em CVPR},  1067--1074.

\bibitem[\protect\citeauthoryear{Shen, Lin, and Huang}{2016}]{shen2016relay}
Shen, L.; Lin, Z.; and Huang, Q.
\newblock 2016.
\newblock Relay backpropagation for effective learning of deep convolutional
  neural networks.
\newblock In {\em ECCV},  467--482.
\newblock Springer.

\bibitem[\protect\citeauthoryear{Simonyan and
  Zisserman}{2014}]{simonyan2014very}
Simonyan, K., and Zisserman, A.
\newblock 2014.
\newblock Very deep convolutional networks for large-scale image recognition.
\newblock {\em arXiv preprint arXiv:1409.1556}.

\bibitem[\protect\citeauthoryear{Sun \bgroup et al\mbox.\egroup
  }{2013}]{patchdeblur_iccp2013}
Sun, L.; Cho, S.; Wang, J.; and Hays, J.
\newblock 2013.
\newblock Edge-based blur kernel estimation using patch priors.
\newblock In {\em ICCP}.

\bibitem[\protect\citeauthoryear{Teh \bgroup et al\mbox.\egroup
  }{2003}]{teh2003energy}
Teh, Y.~W.; Welling, M.; Osindero, S.; and Hinton, G.~E.
\newblock 2003.
\newblock Energy-based models for sparse overcomplete representations.
\newblock {\em JMLR} 4(Dec):1235--1260.

\bibitem[\protect\citeauthoryear{Wang \bgroup et al\mbox.\egroup
  }{2017}]{DBLP:journals/corr/WangLSS17}
Wang, Y.; Liu, R.; Song, X.; and Su, Z.
\newblock 2017.
\newblock An inexact proximal alternating direction method for non-convex and
  non-smooth matrix factorization and beyond.
\newblock {\em arXiv preprint arXiv:1702.08627}.

\bibitem[\protect\citeauthoryear{Xu, Lin, and Zha}{2016}]{xu2016mm}
Xu, C.; Lin, Z.; and Zha, H.
\newblock 2016.
\newblock Relaxed majorization-minimization for non-smooth and non-convex
  optimization.
\newblock In {\em AAAI},  812--818.

\bibitem[\protect\citeauthoryear{Zhao, Mathieu, and
  LeCun}{2016}]{zhao2016energy}
Zhao, J.; Mathieu, M.; and LeCun, Y.
\newblock 2016.
\newblock Energy-based generative adversarial network.
\newblock {\em arXiv preprint arXiv:1609.03126}.

\bibitem[\protect\citeauthoryear{Zoran and Weiss}{2011}]{zoran2011learning}
Zoran, D., and Weiss, Y.
\newblock 2011.
\newblock From learning models of natural image patches to whole image
  restoration.
\newblock In {\em ICCV},  479--486.

\bibitem[\protect\citeauthoryear{Zuo \bgroup et al\mbox.\egroup
  }{2013}]{zuo2013generalized}
Zuo, W.; Meng, D.; Zhang, L.; Feng, X.; and Zhang, D.
\newblock 2013.
\newblock A generalized iterated shrinkage algorithm for non-convex sparse
  coding.
\newblock In {\em ICCV},  217--224.

\end{thebibliography}

\end{document}